\documentclass[10pt,twocolumn,letterpaper]{article}

\usepackage[pagenumbers]{cvpr} 

\usepackage{booktabs}
\usepackage{tabularx}
\usepackage{csquotes}
\usepackage{multirow}
\usepackage[toc,page]{appendix}

\usepackage[dvipsnames]{xcolor}

\newcommand{\quotes}[1]{``#1''}

\newcommand{\ours}{EgoPack\xspace}
\newcommand{\egofourd}{Ego4D\xspace}

\newcommand{\cmark}{\ding{51}}%
\newcommand{\xmark}{\ding{55}}%
\definecolor{up_color}{RGB}{204,51,0}
\definecolor{down_color}{RGB}{202,195,121}

\definecolor{cvprblue}{rgb}{0.21,0.49,0.74}
\usepackage[pagebackref,breaklinks,colorlinks,citecolor=cvprblue]{hyperref}

\usepackage{colortbl}
\usepackage[dvipsnames]{xcolor}
\usepackage{amssymb}
\usepackage{pifont}

\usepackage{pgfplots}
\pgfplotsset{compat=newest}

\usepackage{xcolor-material}
\usetikzlibrary{fit}

\def\paperID{1379} 
\def\confName{CVPR}
\def\confYear{2024}

\title{
A Backpack Full of Skills: \\
Egocentric Video Understanding with Diverse Task Perspectives
}

\author{Simone Alberto Peirone$^{1}$
\quad
Francesca Pistilli$^{1}$
\quad
Antonio Alliegro$^{1,2}$
\quad
Giuseppe Averta$^{1}$
\and
$^{1}$ Politecnico di Torino, $^{2}$ Istituto Italiano di Tecnologia\\
{\tt\small firstname.lastname@polito.it}\\
}

\begin{document}
\maketitle

\begin{abstract}
Human comprehension of a video stream is naturally broad: in a few instants, we are able to understand what is happening, the relevance and relationship of objects, and forecast what will follow in the near future, everything all at once. We believe that - to effectively transfer such an holistic perception to intelligent machines - an important role is played by learning to correlate concepts and to abstract knowledge coming from different tasks, to synergistically exploit them when learning novel skills. To accomplish this, we seek for a unified approach to video understanding which combines shared temporal modelling of human actions with minimal overhead, to support multiple downstream tasks and enable cooperation when learning novel skills. We then propose \ours, a solution that creates a collection of task perspectives that can be carried across downstream tasks and used as a potential source of additional insights, as a backpack of skills that a robot can carry around and use when needed. We demonstrate the effectiveness and efficiency of our approach on four \egofourd benchmarks, outperforming current state-of-the-art methods. Project webpage: \href{https://sapeirone.github.io/EgoPack/}{sapeirone.github.io/EgoPack}.
\end{abstract}    
\vspace{-0.25cm}
\section{Introduction}
\label{sec:intro}
\begin{figure}[t]
    \centering
    \includegraphics[width=0.95\columnwidth]{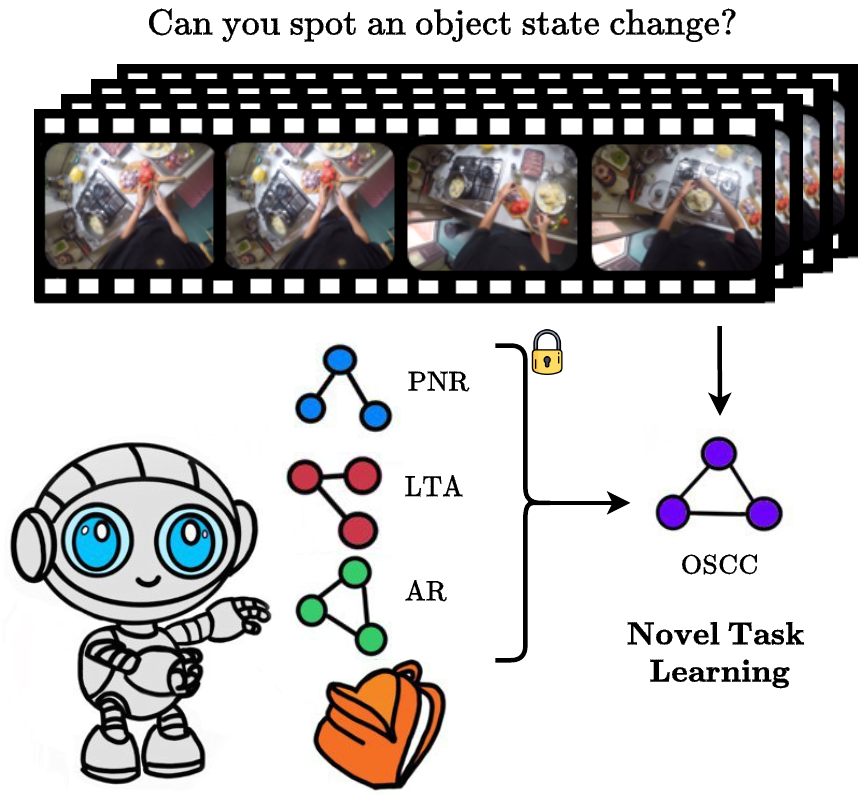}
    \caption{Given a video stream, a robot is asked to learn a novel task, \eg Object State Change Classification (OSCC). 
    To learn the new skill, the robot can access previously gained knowledge regarding different tasks, such as Point of No Return (PNR), Long Term Anticipation (LTA) and Action Recognition (AR), and use it during the learning process to enhance downstream task performance. 
    This knowledge is stored as graphs inside the robot's backpack, always ready to boost a new skill. 
    }
    \vspace{-0.5cm}
    \label{fig:teaser}
\end{figure}
Our daily living activities are extremely complex and diverse, nonetheless humans have the extraordinary ability to reason on the behaviour itself in just a few instants from a visual input. We are able to spot what another person is doing, predict their next actions based on current observations, and understand the implications of an activity, for instance  whether its effects are reversible. 
Observing someone in the kitchen by the worktable, where there is a pack of flour and a jug of water, we can identify that they are a chef kneading flour (reasoning about current activity). We can also forecast that the next step will involve mixing the flour with water (reasoning about the future), and finally obtaining dough (reasoning about implications of these actions). 
This type of holistic reasoning, which is natural for humans, is still a distant goal for artificial intelligence systems. 
The challenge arises not only from the requirement of executing multiple tasks with a single architecture, but also from the necessity of being able to abstract and repurpose such knowledge across-tasks, for example to foster and enhance the learning of novel skills. 
Current research trends in human activity understanding predominantly focus on creating several, hyper-specialised, models. This approach splits the understanding of human activities into distinct skills, with each model being independently trained to rely only on \quotes{task-specific} clues for prediction~\cite{yan2022multiview,zhong2023anticipative,zhang2022actionformer}. However, this approach disregards the valuable insights that could be gleaned from different task perspectives.
A first step in this direction relies on Multi-Task Learning (MTL) to exploit the intuition that knowledge sharing between tasks may improve performance. 
However, the proposed multi-task models have some limitations~\cite{kokkinos2017ubernet}, mostly concerning a negative transfer between tasks, making it difficult to outperform single-task models.
Most importantly, MTL usually assumes the availability of supervision for all tasks at training time, limiting the extension of the models at a later time.

The recently proposed EgoT2 framework~\cite{egot2} offers a unified solution to integrate various egocentric video tasks. It employs an ensemble of diverse, task-specific models and learns to translate task-specific clues through a transformer-based encoder-decoder to benefit one of the tasks. 
Although this approach fosters positive interactions between tasks, it has significant limitations: i) the primary task should be \quotes{known} at training time and present within the task-specific models collection, ii) it necessitates an extensive pretraining process and iii) it lacks a knowledge abstraction, as it relies on task-specific models rather than creating transferable concepts.

Indeed, we argue that an important key to advance the learning capabilities of intelligent systems and to move a step closer to a generalised reasoning on visual understanding involves not only sharing information across tasks, but also abstracting task-specific knowledge for application in new scenarios. 
Considering an ensemble of vision tasks, each offers a distinct perspective on the input stream and extracts different types of information. Our goal is to encapsulate this diverse knowledge to be leveraged in the future to positively impact the learning of a novel skill.
We focus on egocentric video understanding as it is the perfect harbour to study human activities and synergies between tasks. 
There is a strong connection between egocentric tasks. 
For instance, specific actions, like peeling a potato, directly result in a change in the state of the object (the potato in this case), illustrating the interconnected nature of these tasks.

All the above considerations motivate us in investigating new alternatives and we propose a novel framework for knowledge abstraction and sharing called \ours. Our underlying idea, is to exploit a set of known tasks, each one able to interpret an input stream according to its own perspective, to learn reusable knowledge that can aid the learning process of a novel task. 
We show this concept in Fig.~\ref{fig:teaser}, where a robot is equipped with a backpack that figuratively summarises all the knowledge gained from a set of tasks. To learn a new skill, the robot can \quotes{take-out} task-related knowledge from the backpack and leverage it within the learning process. 
The task-specific perspectives are collected in a single pretraining step of a novel multi-task network under the form of prototypes. We exploit a new versatile temporal graph-based architecture shared across all the tasks, with minimal overhead to support each task.

When learning a new skill, \ours promotes the interaction between the different tasks by learning which relevant knowledge to extract from the different perspectives. 
The architecture of \ours is notably flexible, enabling easy adaptation to novel tasks by reusing the previous tasks to facilitate the learning process of any novel task.

We demonstrate the effectiveness and efficiency of our approach on \egofourd~\cite{ego4d}, a large-scale egocentric vision dataset.
To summarise, our main contributions are:
\begin{enumerate}
    \item We present a unified architecture to learn multiple egocentric vision tasks with minimal task-specific overhead;
    \item We introduce \ours, a novel approach that leverages different task perspectives to build a robust knowledge abstraction which can foster the learning of a novel task; 
    \item Our approach outperforms both specialised single and multi-task baselines by leveraging the unique synergies and distinct perspectives of different tasks; 
    \item \ours achieves competitive performance on \egofourd~\cite{ego4d} for all the considered benchmarks, outperforming the state-of-the-art on some.
\end{enumerate}

\section{Related Works}
\label{sec:related_works}

\paragraph{Egocentric Vision}
Egocentric vision captures human activities from the privileged perspective of the camera wearer, allowing a unique point of view on the actions~\cite{betancourt2015evolution,plizzari2023outlook}.
Recently, the field has seen rapid development thanks to the release of several large-scale egocentric vision datasets~\cite{ek55,egtea,epic_tent,ek100,ego4d,sener2022assembly101}.
The rich annotations of these datasets~\cite{ek100,ego4d} allow to tackle a large number of tasks, including action recognition~\cite{nunez2022egocentric}, action anticipation~\cite{furnari2020rulstm,girdhar2021anticipative,zhong2023anticipative}, next active object prediction~\cite{furnari2017next}, action segmentation~\cite{zhang2022actionformer,huang2020improving} and episodic memory~\cite{ramakrishnan2023spotem}.
Previous works in egocentric vision have focused on domain adaptation~\cite{Munro_2020_CVPR,yang2022interact,chen2019temporal,plizzari2023can,planamente2024relative}, multimodal learning~\cite{zehua2033human,gao2020listen, yang2022interact} and large-scale video-language pretraining~\cite{pramanick2023egovlpv2,hiervl,zhao2023learning} to learn better representation for downstream tasks.

\paragraph{Graph Neural Networks for vision tasks}
Traditional neural networks, including Convolutional Neural Networks (CNNs), have been widely used in computer vision, showing impressive performance on a variety of problems~\cite{li2021survey,khan2020survey,gu2018recent}. 
However, these models often assume data lying on a regular domain, such as images that have a grid-like structure. 
In recent years, the interest in developing methods able to provide a more general and powerful type of processing has been growing and particular attention has been given to learning methods on graphs. 
Graph Neural Networks (GNNs) have the innate ability to effectively handle data that lie on irregular domains, such as 3D data~\cite{simonovsky2017dynamic,wang2019dynamic}, robotics~\cite{pistilli2023graph}, molecular chemistry~\cite{kearnes2016molecular}, and social or financial networks~\cite{fan2019graph}, and to model complex data relations~\cite{sanchez2020learning}. 
Recently, transformer-based architectures had a great impact on vision application. 
Despite Transformers and GNNs share some similarities in their ability to handle various data types, they are fundamentally different in their core architectures and the specific ways they process data. GNNs can model the topology of a graph and the relations between nodes while also inheriting all the desirable properties of classic convolutions: locality, hierarchical structures and efficient weight reuse.
In video understanding GNNs have been applied to action localisation~\cite{huang2020improving,zeng2019graph,ghosh2020stacked,rashid2020action}, to build a knowledge graph from human actions~\cite{ghosh2020all}, to model human-object interactions~\cite{dessalene2020egocentric, dessalene2021forecasting} or to build a topological map of the environment~\cite{nagarajan2020ego}.

\paragraph{Multi-Task Learning}
MTL~\cite{caruana1997multitask, zhang2021survey} tackles the problem of learning to solve multiple task simultaneously. 
The development of this strategy is justified by the intuition that complex settings require solving multiple tasks, for instance autonomous driving~\cite{Huang_2023_ICCV}, robotics and natural language processing. 
Furthermore, these networks can bring the theoretical advantage of sharing complementary information to improve performance. 
Several works have been done in this direction~\cite{kokkinos2017ubernet, huang2020mutual, fifty2021efficiently, chen2022unified, Chen_2023_ICCV, shi2023deep, Huang_2023_ICCV, ci2023unihcp}, focusing on which parameters or tasks is better to share~\cite{kang2011learning, guo2020learning, standley2020tasks, sun2020adashare} and promoting synergies between tasks~\cite{kapidis2019multitask, wang2021interactive}. 
Such methods encounter the problem of negative transfer~\cite{kokkinos2017ubernet} and sharing with unrelated tasks~\cite{guo2020learning, standley2020tasks} consequently suffering of task competition and not being able to benefit from information sharing between tasks. To overcome these limitations, several methods have been proposed to balance task-related losses~\cite{kendall2018multi, chen2018gradnorm, sinha2018gradient}, to dynamically prioritise tasks~\cite{guo2018dynamic}, to reduce gradient interference between tasks~\cite{gradient_surgery} or to exploit task interactions at multiple scales~\cite{vandenhende2020mti}. 
Unfortunately, all these solutions require extensive task-specific tuning, and are not able to build an holistic perception across tasks.
Few works have explored MTL in the field of egocentric vision~\cite{kapidis2019multitask,egot2,huang2020mutual,wang2021interactive}.
Among these, the recently proposed EgoT2~\cite{egot2} builds an ensemble of diverse, task-specific models. 
The features of the different models are projected into a common feature space and processed through a transformer-based encoder-decoder to translate the contributions of different tasks and generate predictions for the primary task. Notably, the primary task has to be part of the task-specific models. This approach fosters positive interactions between tasks, resulting in improved performance compared to the single-task models. 
However, it has some limitations, as it is not able to build knowledge abstractions that can be easily transferred to novel tasks.
Instead, we propose a model that can build a robust backpack of task perspectives that can be used in learning any novel tasks.

\section{Method}
\label{sec:method}

\begin{figure*}[htbp]
    \centering
    \includegraphics[width=0.825\textwidth,trim=0in 0.35in 1.25in 0.4in]{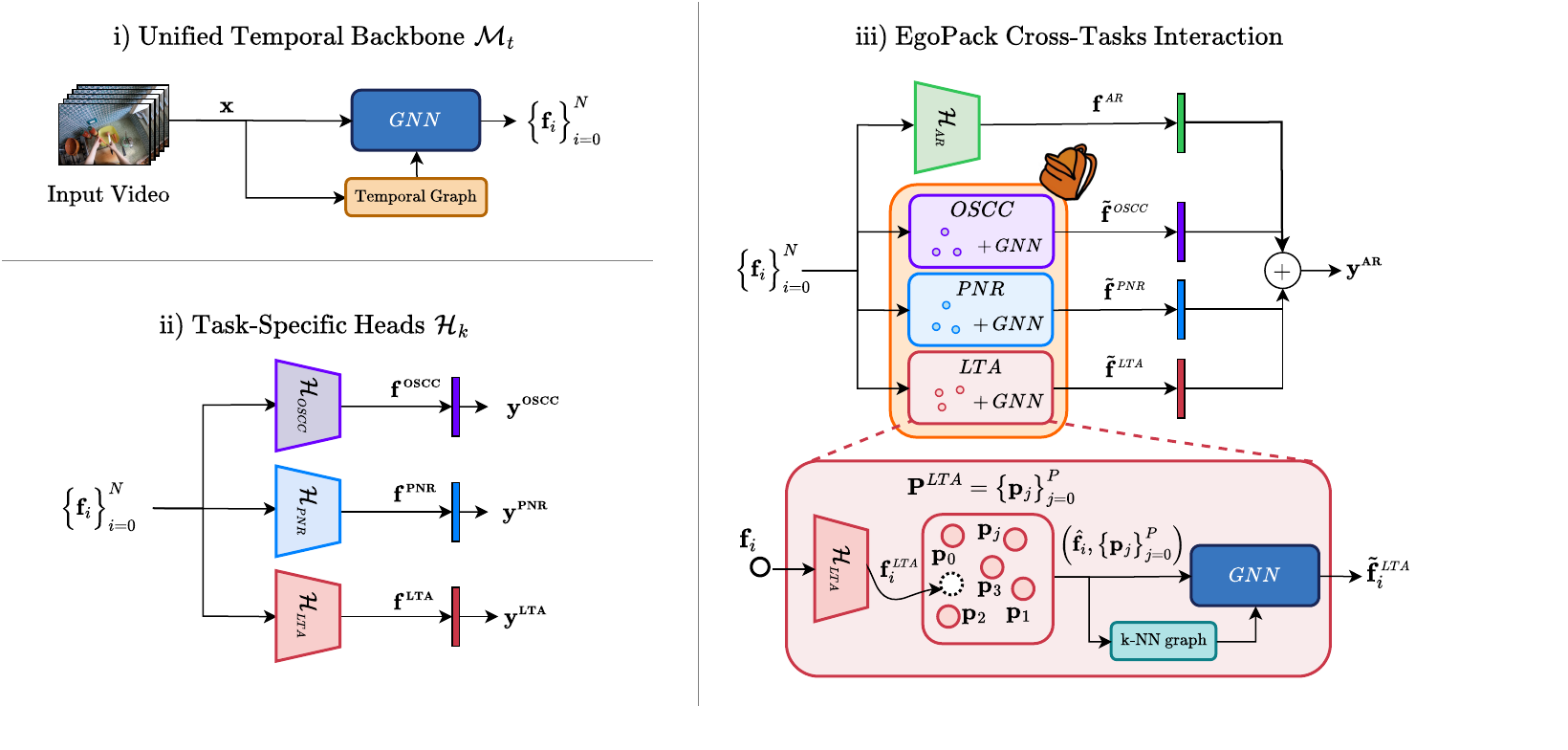}
    \caption{
    Architecture of \ours when Action Recognition (AR) is the novel task.
    Videos are interpreted as a graph, whose nodes $\mathbf{x}_i$ represent actions, encoded as features, and edges connect temporally close segments. This representation enables the design of a \emph{Unified Temporal Backbone} to learn multiple tasks with a shared architecture and minimal \emph{Task-Specific Heads}, leveraging GNNs for temporal modelling.
    We exploit this architecture to jointly learn $K$ tasks, \eg OSCC, LTA and PNR. After this training process, we extract a set of prototypes $\mathbf{P}^k$ that summarise what the network has learnt from each task $\mathcal{T}_k$, like a backpack of skills that we can carry over.
    In this \emph{Cross-Tasks Interaction} phase, the network can peek at these different task-perspective to enrich the learning of the novel task.
}
\vspace{-2.5mm}
    \label{fig:architecture}
\end{figure*}

We tackle a task cooperation setting, in which an egocentric vision model is able to exploit previously acquired knowledge over task perspectives to foster the learning process of any novel task.
We formulate the proposed setting in Sec.~\ref{sec:method_setting}.
We present a unified temporal architecture to model multiple tasks in Sec.~\ref{sec:method_mtl}, a key step to enable knowledge sharing between tasks.
Finally, Sec.~\ref{sec:method_adapt} presents our novel approach \ours to enable efficient transfer of different task perspectives to novel tasks.

\subsection{Setting}\label{sec:method_setting}
A task $\mathcal{T}$ is associated with a dataset $\mathcal{D} = \{(v_{i}, y_{i})\}_{i=1}^{N}$, where $v_{i}$ is a video segment of arbitrary length, $y_{i}$ is the associated ground truth label and $N$ is the number of segments.
Our approach follows a two-stages training process.
First, a model $\mathcal{M}$ is trained on a set of $K$ tasks $\{\mathcal{T}_0,\dots,\mathcal{T}_K\}$, under a Multi-Task Learning framework with hard-parameter sharing~\cite{ruder2017overview} to encourage the model to learn more general and task-agnostic representations thanks to the joint supervision of multiple tasks.
Then, the model is presented with a new task $\mathcal{T}_{K+1}$ to learn, without access to the supervision of the previous tasks.
In this scenario, the new task may benefit from potential semantic affinities with the previously seen tasks. 
For example, a model that has learnt to detect object changes may apply this knowledge for action recognition and vice-versa, as some actions are associated with object changes, \eg \emph{cutting something}, while others are not, \eg \emph{moving an object}. 
Our goal is to make these semantic affinities more explicit (and exploitable) so that the new task can learn to repurpose these \emph{perspectives} from previous tasks to improve performance, a step towards more holistic models that seamlessly share knowledge between tasks.

\subsection{A Unified Architecture for Egocentric Tasks}
\label{sec:method_mtl}
The main premise of our method is that different egocentric vision tasks can be modelled using a shared architecture with minimal differences between tasks. 
Under this assumption, videos can be seen as a sequence of $N$ temporal segments encoded as $\mathbf{x} = \{\mathbf{x}_1, \mathbf{x}_2, \dots, \mathbf{x}_N \}$, where $\mathbf{x}_i \in \mathbb{R}^D$ represents the $D$-dimensional features of segment $v_i$ extracted using some video features extractor $\mathcal{F}$, \eg SlowFast~\cite{slowfast} or Omnivore~\cite{omnivore}.
Such sequence could be interpreted as a temporal graph $\mathcal{G}(\mathcal{X}, \mathcal{E})$, whose nodes $\mathbf{x}_i \in \mathcal{X}$ represent the segments of the video, and edges $e_{ij} \in \mathcal{E}$ connect nodes $\mathbf{x}_i$ and $\mathbf{x}_j$ with a temporal distance considered relevant when lower than a threshold $\tau$.
The connectivity of the graph defines the extent of its temporal modelling, \ie connecting further apart nodes enables longer range temporal understanding which could benefit for example anticipation tasks.
The threshold $\tau$ depends on the task at hand and more implementation details are provided in Sec.~\ref{sec:experiments_impl_details}.
The temporal position of each node in the sequence is encoded by adding to the node embeddings a positional encoding~\cite{vaswani2017attention}.

This formulation enables the use of Graph Neural Networks (GNNs) to learn the complex temporal relations between video segments and to cast different egocentric vision tasks as graph prediction tasks, such as node-level or graph-level classification, as shown in Fig.~\ref{fig:egocentric_graph_tasks}. 
This assumption is reflected in our approach by decomposing the multi-task model $\mathcal{M}$ into two components: a general \emph{temporal} backbone $\mathcal{M}_{t}: \mathbb{R}^D \mapsto \mathbb{R}^{D_t}$, and a set of task-specific projection heads $\mathcal{H}_k: \mathbb{R}^{D_t} \mapsto \mathbb{R}^{D_k}$ mapping the graph and/or the nodes to the desired output space for task $\mathcal{T}_k$ with dimension $D_k$, as shown in Fig.~\ref{fig:architecture}.
$\mathcal{M}_{t}$ is a GNN with $L$ layers that takes as input the temporal sequence $\mathbf{x}$ and provides as output the updated feature vectors $\mathbf{f}=\{\mathbf{f}_1, \mathbf{f}_2, \dots , \mathbf{f}_N\}$. 
At layer $l$, node embeddings are projected and combined with their neighbours, following the GraphSAGE architecture~\cite{graphsage}:
\begin{equation}
    \small
    \mathbf{f}^{(l+1)}_i = \mathbf{W}^{(l)}_{r} \mathbf{f}_i^{(l)} + \mathbf{W}^{(l)} \cdot \mathbf{g}^{(l+1)}_i + \mathbf{b}^{(l)} ,
\end{equation}
where $\mathbf{f}_i^{(l)} \in \mathbb{R}^{D^{(l)}_t}$ are the features of node $\mathbf{x}_i$, $\mathbf{b}^{(l)} \in \mathbb{R}^{D^{(l+1)}_t}$ is a bias term, $\mathbf{W}^{(l)}_{r},\mathbf{W}^{(l)} \in \mathbb{R}^{D^{(l+1)}_t \times D^{(l)}_t}$ are the weight matrices associated to the root node and the aggregated neighbours' contribution $\mathbf{g}^{(l+1)}_i$ respectively. 
The latter is computed as:
\begin{equation}
    \mathbf{g}^{(l+1)}_i = \underset{\rm \mathbf{f}_j \, \in \, \mathcal{N}_i}{mean} \Big(\phi\big(\mathbf{W}^{(l)}_{p}\mathbf{f}_j^{(l)} + \mathbf{b}^{(l)}_p\big)\Big) , \\
\end{equation}
where $\mathbf{W}^{(l)}_{p} \in \mathbb{R}^{D^{(l)}_t \times D^{(l)}_t}$ projects the neighbours before the aggregation step, $\phi$ is a non-linearity, $\mathbf{b}_p^{(l)} \in \mathbb{R}^{D^{(l+1)}_t}$ is a bias term and $\mathcal{N}_i$ is the set of neighbours of node $\mathbf{x}_i$.
Each layer is followed by Layer Normalization~\cite{layernorm} and a LeakyReLU activation function. A residual connection around the temporal GNN allows the network to preserve the original features. 
Intuitively, the neighbourhood $\mathcal{N}_i$ reflects the temporal dependencies of the input sequence and the GNN allows to iteratively extend the temporal receptive field of each node. 

\vspace{-2.5mm}
\paragraph{Task-specific heads}
The output features of the temporal backbone $\mathcal{M}_{t}$ are shared across the different downstream tasks. 
To project these features into task-specific components, we add a set of projection heads $\mathcal{H}_k$, one for each task $\mathcal{T}_k$.
For graph classification tasks, the nodes of each graph are aggregated using max pooling to obtain a unique features representation.
In each head, a MultiLayer Perceptron outputs the task-specific features $\mathbf{f}_{i}^k \in \mathbb{R}^{D^k}$ and is followed by a linear layer to compute the task logits $\mathbf{y}_{i}^k \in \mathbb{R}^{D^k_o}$, where $D^k_o$ is the number of labels for task $\mathcal{T}_k$. 
By limiting the task-specific portion of the network to the heads while sharing the temporal backbone, we can obtain the perspective of all tasks with a single forward through the latter.
The network is trained on all the tasks by averaging their losses.

\subsection{Learning a novel task with a backpack}
To solve the new task, the naive approach would be to finetune the model, adding a new head $\mathcal{H}_{K+1}$ and possibly updating the temporal backbone $\mathcal{M}_{t}$.
However, finetuning may not fully leverage the insights from other tasks as it could result in the loss of the previously acquired knowledge, as confirmed experimentally in Sec.~\ref{sec:experiments_quantitative}.

\paragraph{Building the backpack}
We propose to explicitly model the perspectives of the different tasks as a set of task-specific prototypes that abstract the knowledge gained by the previously seen tasks and can be accessed by novel tasks. We call this approach \ours and provide an overview in Fig.~\ref{fig:architecture}. 
These task-specific prototypes are collected from videos annotated for action recognition, as human actions can be seen as the common thread behind the different tasks.
Practically, we forward these samples through the temporal backbone and take the output of the different task-specific projection heads, thus encoding the perspective of each task given the same input video.
Finally, the features obtained from each task are aggregated according to the \textit{verb} and \textit{noun} labels of the action, effectively summarising the perspective of each task given the same input action.
The result is a set of prototypes $\mathbf{P}^k = \{ \mathbf{p}_0^k, \mathbf{p}_2^k, \dots, \mathbf{p}_P^k \}$ $ \in \mathbb{R}^{P \times D_k}$ for each task $\mathcal{T}_k$, where $P$ is the number of unique \textit{(verb, noun)} pairs in the dataset and $D_k$ is the size of the task-specific features.
These prototypes are frozen and represent a \quotes{summary} of what the models has learnt during the multi-task pretraining process, creating an abstraction of the gained knowledge. They can be then reused when learning novel tasks, like a backpack of skills that the model can carry over.

\paragraph{Leveraging the backpack}
\label{sec:method_adapt}
During the learning process of the novel task $\mathcal{T}_{K+1}$, the model can exploit the task prototypes obtained via the task-specific heads.
As before, the output of the temporal backbone $\mathbf{f}_i$ is forwarded through all the projection heads to obtain the task-specific features $\mathbf{f}^{k}_i$.
These features are used as \emph{queries} to match the corresponding task prototypes $\mathbf{P}^k$, selecting the $k$-Nearest Neighbours among the prototypes using cosine similarity in the features space. 
Task features and their neighbouring prototypes form a \emph{graph-like} structure, on which message passing can be used to enrich the task-specific features $\mathbf{f}_{i}^k$, following an iterative refinement approach. In particular, at each layer $l$ we select the closest prototypes with $k$-NN and update the features $\mathbf{f}_{i}^{(l), k}$ according to the following rule:
\begin{equation}
    \small
    \mathbf{f}^{(l+1), k}_{i} = \mathbf{W}^{(l)}_{r} \mathbf{f}_{i}^{(l), k} + \mathbf{W}^{(l)} \cdot  \underset{\mathbf{p}^k_{j} \, \in \, \mathcal{N}_i^{(l),k}}{max} \mathbf{p}^k_{j},
\end{equation}
where $\mathbf{p}^k_{j} \, \in \, \mathcal{N}_i^{(l),k}$ are the closest prototypes in $\mathbf{P}^k$ to $\mathbf{f}_{i}^{(l), k}$ and $\mathbf{W}^{(l)}_{r},\mathbf{W}^{(l)} \in \mathbb{R}^{D^k \times D^k}$ are the weight matrices associated to the input features and the aggregated neighbours respectively.
Notably, only the task features are updated while the task prototypes remain frozen to preserve the original perspectives seen by the network.

In this process, the task-specific heads $\mathcal{H}_k$ are initialised from the multi-task training and possibly updated during the task-specific finetuning process, allowing the model to freely explore the set of task prototypes and to select the most informative ones for each input sample.
After the interaction phase, the refined features $\tilde{\mathbf{f}}_{i}^k$ are fed to a classifier module to obtain the task logits $\mathbf{y}_{i}^{k} \in \mathbb{R}^{D^k_o}$ for each task $\mathcal{T}_k$ in the backpack.
The final prediction is the sum of the pre-softmax logits coming from the different tasks and the output of a new head $\mathcal{H}_{K+1}$ for the novel task.
Intuitively, we allow each task to cast a vote on the final prediction, based on its perspective on the same video segment.
In this phase, the temporal network, the task-specific heads and the weights of the GNNs are trained jointly using only the supervision of the novel task $\mathcal{T}_{K+1}$.

\section{Experiments}
\label{sec:experiments}
We evaluate \ours on four Ego4d Human-Object Interaction benchmarks: Action Recognition (AR)\footnote{AR is not an official \egofourd benchmark and was derived from the LTA annotations by~\cite{egot2}. To be consistent with previous works, we use the v1 version of the LTA annotations.}, Long Term Action Anticipation (LTA), Object State Change Classification (OSCC) and Point Of No Return (PNR).
We report verb and noun top-1 accuracy for AR, accuracy for OSCC, edit distance for LTA and temporal localisation error (in seconds) for PNR.

\subsection{Implementation Details}
\begin{figure*}[htbp]
    \centering
    \hfill
    \begin{subfigure}[b]{0.3\textwidth}
        \centering
        \includegraphics[width=\textwidth]{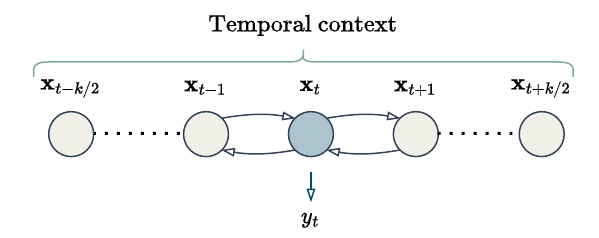}
        \caption{Node classification (AR, PNR)}
        \label{fig:egocentric_graph_tasks_AR}
    \end{subfigure}
    \hfill
    \begin{subfigure}[b]{0.3\textwidth}
        \centering
        \includegraphics[width=\textwidth]{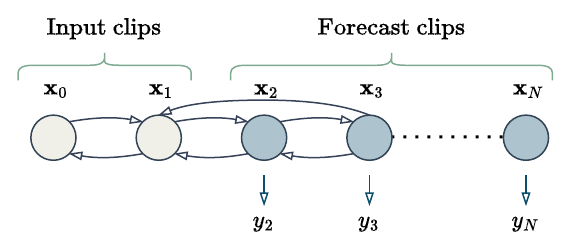}
        \caption{Future node classification (LTA)}
        \label{fig:egocentric_graph_tasks_LTA}
    \end{subfigure}
    \hfill
    \begin{subfigure}[b]{0.3\textwidth}
        \centering
        \includegraphics[width=\textwidth]{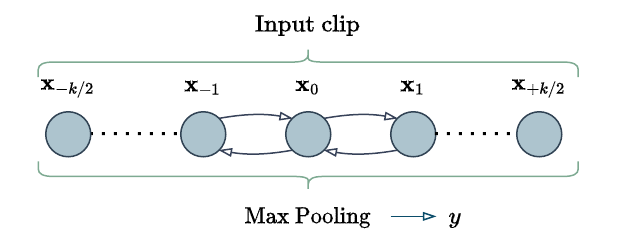}
        \caption{Graph classification (OSCC)}
        \label{fig:egocentric_graph_tasks_OSCC}
    \end{subfigure}
    \hfill
    \caption{Egocentric vision tasks as graph prediction tasks. In AR and LTA, each node is an action within a temporal sequence and the objective is to predict the verb and noun labels of the nodes. 
    In OSCC and PNR, nodes represent different temporal segments of the video clip and the goal is to output a global prediction for the entire graph (OSCC) or the individual nodes (PNR).
    }
    \vspace{-3.5mm}
    \label{fig:egocentric_graph_tasks}
\end{figure*}
\label{sec:experiments_impl_details}
We use Omnivore Swin-L~\cite{omnivore} features pre-trained on Kinetics-400~\cite{quo_vadis}, released as part of \egofourd~\cite{ego4d} and extracted using dense sampling over a window of 32 frames with a stride of 16 frames and features size 1536.
In principle, \ours is agnostic to the underlying features extractor and could adopt other architectures.
Following previous works on \egofourd~\cite{plizzari2023can} we use TRN~\cite{TRN} to temporally aggregate features from the three sub-segments of each input sample.
The mapping between videos of each task and its corresponding temporal graph is task dependent, as shown in Fig.~\ref{fig:egocentric_graph_tasks}:
\begin{itemize}
    \item \textbf{Action Recognition (AR)}: actions are mapped to the nodes of the temporal graph $\mathcal{G}$, and edges connect each node to the previous and next (Fig.~\ref{fig:egocentric_graph_tasks_AR}). To account for the variable length of videos, actions are processed in fixed size windows.
    \item \textbf{Long Term Anticipation (LTA)}: each input clip is mapped to a node in $\mathcal{G}$. Then, a sequence of new nodes is added to the graph, equivalent in number to the clips to forecast. These nodes are initialised with the mean features of the input clips and are connected to the previous and subsequent nodes in the sequence, as well as to the input clips (Fig.~\ref{fig:egocentric_graph_tasks_LTA}).
    \item \textbf{Object State Change Classification (OSCC) and Point of No Return (PNR)}: each input segment is further split into $n$ sub-segments to account for the finer temporal granularity required by these tasks. Each sub-segment is mapped to a node in $\mathcal{G}$, and edges connect each node to the previous and next (Fig.~\ref{fig:egocentric_graph_tasks_OSCC}). 
\end{itemize} 
Tasks have different annotations and are modelled as separate graphs, even though the temporal model is shared.
The task prototypes are built using samples from the train split of the AR dataset.
Tasks are trained with standard cross entropy loss, with the exception of PNR which uses binary cross entropy.
\ours is trained for 30, 40 and 10 epochs for AR, LTA and OSCC/PNR respectively, with a learning rate of $1e-4$ and a linear warm-up for the first 5 epochs, using Adam optimiser and batch size 16.
All tasks share the same temporal and cross-task interaction architecture, with minimal task-specific hyper-parameter tuning. More details are reported in the supplementary.

\subsection{Quantitative results}
\label{sec:experiments_quantitative}
\begin{table*}[ht]
\centering
\footnotesize
\setlength{\tabcolsep}{5.3pt}
\begin{tabular}{lcccccccc}
\toprule

& \multirow{3}{*}{\textbf{\shortstack{Trained on\\frozen features}}} & \multicolumn{2}{c}{\textbf{AR}} & \multicolumn{1}{c}{\textbf{OSCC}} & \multicolumn{2}{c}{\textbf{LTA}} & \multicolumn{1}{c}{\textbf{PNR}} \\
\cmidrule(lr){3-8}

& & \textbf{Verbs Top-1 (\%)} & \textbf{Nouns Top-1 (\%)} & \textbf{Acc. (\%)} & \textbf{Verbs ED ($\downarrow$)} & \textbf{Nouns ED ($\downarrow$)} & \textbf{Loc. Err. (s) ($\downarrow$)}\\

\midrule

Ego4D Baselines~\cite{ego4d} & \xmark & 22.18 & 21.55 & 68.22 & 0.746 & 0.789 & 0.62 \\
EgoT2s~\cite{egot2} & \xmark & 23.04 & 23.28 & \textbf{72.69} & 0.731 & 0.769 & \textbf{0.61} \\

\midrule
\midrule

MLP & \cmark & 24.08 & 30.45 & 70.47 & 0.763 & \textbf{0.742} & 1.76 \\
Temporal Graph & \cmark & 24.25 & 30.43 & 71.26 & 0.754 & 0.752 & \textbf{0.61} \\

Multi-Task Learning & \cmark & 22.05 & 29.44 & 71.10 & 0.740 & 0.746 & 0.62 \\
Task Translation$^\dagger$ & \cmark & 23.68 & 28.28 & 71.48 & 0.740 & 0.756 & \textbf{0.61} \\

\midrule
\ours & \cmark & \textbf{25.10} & \textbf{31.10} & 71.83 & \textbf{0.728} & 0.752 & \textbf{0.61} \\
\bottomrule

\end{tabular}
\caption{\ours on \egofourd HOI tasks. MLP is a simple baseline consisting of a few linear layers, while Temporal Graph models all tasks using a unified temporal graph-based architecture. MTL~\cite{ruder2017overview} uses hard parameter sharing to jointly learn all tasks, which may result in negative transfers. Ego-T2s~\cite{egot2} learns to translate features across tasks to optimise the primary task. \ours builds on the unified architecture of the Temporal Graph and learns to exploit the perspective of different tasks for efficient transfers to the novel task. Performances of \ours are evaluated over three runs using accuracy for AR and OSCC, Edit Distance for LTA and temporal localisation error for PNR. $^\dagger$Our implementation of the task translation mechanism from EgoT2~\cite{egot2} using Omnivore features.}
\vspace{-.25cm}
\label{tab:main_results}
\end{table*}
We show the main results of \ours in Table~\ref{tab:main_results}. 
To assess the validity of our approach, we proceed incrementally starting from single tasks models, \ie each task is trained separately. 
In this setting, we compare a simple MLP baseline trained on the temporally aggregated features against our temporal graph methodology, which exhibits superior average performance. The improvement is particularly evident in the PNR task, \eg from $1.76\,s$ to $0.61\,s$, where the subpar outcomes of the MLP can be attributed to the lack of explicit temporal modelling. 
In addition to higher performance, the temporal graph enables all the tasks to be modelled using a unified architecture which allows to train all the tasks at the same time (MTL).
With the MTL model, we observe a significant drop in average performance, mostly driven by worse accuracy in AR and OSCC.
This behaviour is the result of negative transfers between tasks when they are trained together~\cite{Wu2020Understanding}.

\paragraph{Cross-Task Interactions}
We compare our approach \ours for efficient cross task interaction with EgoT2~\cite{egot2}, which learns to combine multiple task-specific frozen models to solve one of them.
Unlike \ours, the learning process of EgoT2 is divided in two stages, \ie a pre-training step where each individual task is learned from scratch and a task-specific translation step, where just one task of the collection is fine-tuned. Notably, both steps require the supervision of the downstream task.
On the contrary, the multi-task pre-training of \ours is agnostic to the novel downstream task, potentially allowing to transfer the gained knowledge to any new task.
To ensure a fair comparison with \ours, we re-implemented the task translation mechanism proposed by EgoT2 on top of our \textit{Temporal Graph} single task models using Omnivore features.
This approach is indicated as \textit{Task Translation} in Table~\ref{tab:main_results}.
Additional details on its implementation are provided in the supplementary.
One of the main benefit of our approach is that it requires a single forward pass through the features extraction and temporal backbones to obtain the perspectives of the different tasks, unlike EgoT2 which requires a forward pass for each single task model.
Notably, we also highlight that \ours obtains better or comparable performance even though the backbone used for features extraction was not trained on \egofourd.

\paragraph{Ablation of the different contributions}
\begin{table}[htbp]
    \centering
    \footnotesize
    \setlength{\tabcolsep}{5.4pt}
    \begin{tabular}{l|ccc|c}
        \toprule
         & \multirow{2}{*}{\textbf{\shortstack{Temp.\\model}}} & \multirow{2}{*}{\textbf{\shortstack{Multi-Task\\Objective}}} & \multirow{2}{*}{\textbf{\shortstack{Cross-Task\\Interaction}}} & \multirow{2}{*}{\textbf{\shortstack{Metrics\\Average ($\Delta$)}}} \\
         & & & &\\
         \midrule
         MLP & \xmark & \xmark & \xmark & \scriptsize{0.416}  \\
         Temp. & \cmark & \xmark & \xmark & \scriptsize{0.433 ({\scriptsize +4.22\%})}\\
         \midrule
         Task Transl. & \cmark & \xmark & \cmark & \scriptsize{0.431 ({\scriptsize +3.61\%})}\\
         MTL & \cmark & \cmark & \cmark & \scriptsize{0.430 ({\scriptsize +3.50\%})}\\
         MTL+FT & \cmark & \cmark & \cmark & \scriptsize{0.437 ({\scriptsize +5.02\%})}\\
         \ours & \cmark & \xmark & \cmark & \scriptsize{\textbf{0.441} ({\scriptsize \textbf{+6.10\%}})} \\
         \bottomrule
    \end{tabular}
    \caption{Ablation of the different contributions in \ours, measured according an aggregated score, computed as the mean of the standardised metrics across tasks.}
    \vspace{-5mm}
    \label{tab:ablation_tasknomy}
\end{table}
We summarise the main steps leading to \ours in Table~\ref{tab:ablation_tasknomy}, using an aggregated metric to capture the overall improvement across the various tasks when compared to the baseline. 
The metric is computed as an average of the individual task metrics.
We adjusted the metrics by taking one minus the score for LTA and PNR, as lower values are preferable, and clipped the PNR localisation error at 1.0 to have the same scale across all the metrics.
Temporal modelling alone greatly improves the score compared to the baseline.
Although MTL allows to train under a multi-task objective, it clearly under-performs the temporal model due to negative transfers~\cite{kokkinos2017ubernet}. \emph{Task Translation} partially recovers this gap on some tasks as shown in Table~\ref{tab:main_results}, but overall the aggregated metric is comparable with MTL.
We speculate that the marginal improvement of \textit{Task Translation} compared to MTL lies in the limited task-specific context the former has access to, as it can peek at the different perspectives of the auxiliary tasks only for the input video at hand, rather than looking at the entire knowledge gained by the model.
On the contrary, the task prototypes of \ours allow to carry a more complete summary of what the models has learnt from which it can extract useful knowledge based on the sample and the task at hand.
To validate that the benefits of \ours were not brought by the MTL pre-training alone, we also introduce a \emph{MTL+FT} baseline where a new task-specific head is finetuned for the novel task, without access to the output of the other heads. The limited performance of this configuration could be explained by the model losing the knowledge learnt during the multi-task learning, without a significant improvement over the single-task baselines, thus only partially reusing the gained knowledge.
On the contrary, \ours preserves this knowledge in the form of prototypes, which proves to be effective for retaining the model's knowledge when learning a new task.

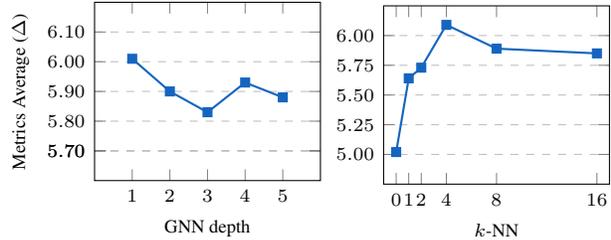
\begin{figure}
    \centering
    \hspace{-6mm}
    \begin{subfigure}[b]{0.55\columnwidth}
        \centering
        \centering
\begin{tikzpicture}
\begin{axis}[
    enlargelimits=false,
    ylabel={Metrics Average ($\Delta$)},
    xlabel={GNN depth},
    xmin=0, xmax=6,
    ymin=5.6, ymax=6.20,
    xtick={1,2,3,4,5},
    ytick={5.70,5.70,5.80,5.90,6.00,6.10},
    ymajorgrids=true,
    grid style=dashed,
    width=\columnwidth,
    legend columns=-1,
    legend style={draw=none},
    every axis plot/.append style={thick},
    label style={font=\scriptsize},
    tick label style={font=\scriptsize},
    yticklabel style={
        /pgf/number format/.cd,
        fixed,
        fixed zerofill,
        precision=2
    }
]
\addplot[color=MaterialBlue800,mark=square*,mark size=1.5pt]table[x index=0,y index=1,col sep=comma]{sec/figures/ablations/data/depth.txt};
\end{axis}
\end{tikzpicture}
    
    \end{subfigure}
    \hspace{-6mm}
    \begin{subfigure}[b]{0.55\columnwidth}
        \centering
        \centering
\begin{tikzpicture}
\begin{axis}[
    enlargelimits=false,
    xlabel={$k$-NN},
    xmin=-1, xmax=17,
    ymin=4.75, ymax=6.25,
    xtick={0,1,2,4,8,16},
    ytick={5.00,5.25,5.50,5.75,6.00},
    ymajorgrids=true,
    grid style=dashed,
    width=\columnwidth,
    legend columns=-1,
    legend style={draw=none},
    every axis plot/.append style={thick},
    label style={font=\scriptsize},
    tick label style={font=\scriptsize},
    yticklabel style={
        /pgf/number format/.cd,
        fixed,
        fixed zerofill,
        precision=2
    }
]
\addplot[color=MaterialBlue800,mark=square*,mark size=1.5pt]table[x index=0,y index=1,col sep=comma]{sec/figures/ablations/data/k.txt};
\end{axis}
\end{tikzpicture}
    \end{subfigure}
    \caption{
    Parameter analysis for the cross-tasks interaction module of \ours.
    We analyse the impact on performance of GNN depth and the number of nearest neighbours denoted as $k$-NN. }
    \vspace{-.5cm}
    \label{fig:ablation}
\end{figure}

\vspace{-1mm}
\paragraph{Depth of the GNN and the selection of $k$}
We observe that \ours is quite robust to the number of GNN layers in the interaction stage between the input features and the task prototypes, as shown in Fig.~\ref{fig:ablation}.
Regarding the selection of the $k$ parameter, we compare the \emph{MTL+FT} baseline ($k=0$) with \ours. The best performance is achieved at $k=4$ with a saturating trend afterwards, showing that interacting with a limited number of prototypes is sufficient.

\vspace{-1mm}
\paragraph{Results on the test set}
We compare \ours on the test set of PNR, OSCC and LTA benchmarks, to validate the improvements and soundness of \ours.
In this setting, a fair comparison between methods is challenging because of the use of different backbones, supervision levels, ensemble strategies and challenge-specific tuning, such as training also on the validation set.
Remarkably, we achieve SOTA performances in LTA, outperforming the other methods that finetune the entire backbone, with a more evident benefit in the verbs edit distance.
In PNR, we closely match other approaches while the improvement is more limited in the OSCC task.
In this task, we notice a relevant impact of the \egofourd pretraining on the performance.
We provide a more in-depth description of the differences between these methods in the supplementary materials.

\begin{table}[ht]
\centering
\footnotesize
\setlength{\tabcolsep}{3pt}
\begin{tabular}{lcccc}

\toprule

\textbf{PNR} & \textbf{Ego4D Pt.} & \multicolumn{3}{c}{\textbf{Loc. Error (s) ($\downarrow$)}}\\

\midrule

CNN LSTM~\cite{ego4d} & \xmark & \multicolumn{3}{c}{0.76} \\

EgoVLP~\cite{lin2022egocentric} & \cmark & \multicolumn{3}{c}{0.67} \\

EgoT2~\cite{egot2} & \xmark & \multicolumn{3}{c}{0.66} \\

\ours & \xmark & \multicolumn{3}{c}{\textbf{0.66}} \\

\midrule
\midrule

\textbf{OSCC} & \textbf{Ego4D Pt.} & \multicolumn{3}{c}{\textbf{Accuracy (\%)}}\\

\midrule

I3D RN-50~\cite{ego4d} & \xmark & \multicolumn{3}{c}{67.6} \\

EgoVLP~\cite{lin2022egocentric} & \cmark & \multicolumn{3}{c}{74.0} \\

EgoT2 (EgoVLP)~\cite{egot2} & \cmark & \multicolumn{3}{c}{\textbf{75.0}} \\
EgoT2 (I3D)~\cite{egot2} & \xmark & \multicolumn{3}{c}{71.0} \\

\ours & \xmark & \multicolumn{3}{c}{72.1} \\

\midrule
\midrule

\textbf{LTA} & \textbf{Ego4D Pt.} & \textbf{Verb ($\downarrow$)} & \textbf{Noun ($\downarrow$)} & \textbf{Action ($\downarrow$)} \\

\midrule

SlowFast~\cite{ego4d} & \xmark & 0.739 & 0.780 & 0.943 \\

EgoT2~\cite{egot2} & \xmark & 0.722 & 0.764 & 0.935 \\

HierVL~\cite{hiervl} & \cmark & 0.724 & \textbf{0.735} & 0.928 \\

I-CVAE~\cite{mascaro2023intention} & \xmark & 0.741 & 0.740 & 0.930 \\

\ours & \xmark & \textbf{0.721} & \textbf{0.735} & \textbf{0.925} \\

\bottomrule

\end{tabular}
\caption{
Comparison of \ours on the test set of the Ego4D benchmarks. 
For a fair comparison, we differentiate between methods that have been pretrained on full \egofourd (\cmark) and those that have been trained only on the benchmark data (\xmark), which includes \ours.
}
\vspace{-2.5mm}
\label{tab:challenge}
\end{table}

\subsection{Qualitative results}
\paragraph{Closest Task Prototypes}
\begin{figure}[htbp]
    \centering
    \begin{subfigure}{\columnwidth}
        \includegraphics[width=0.95\columnwidth, trim=0.25cm 0.8cm 0.25cm 0.25cm]{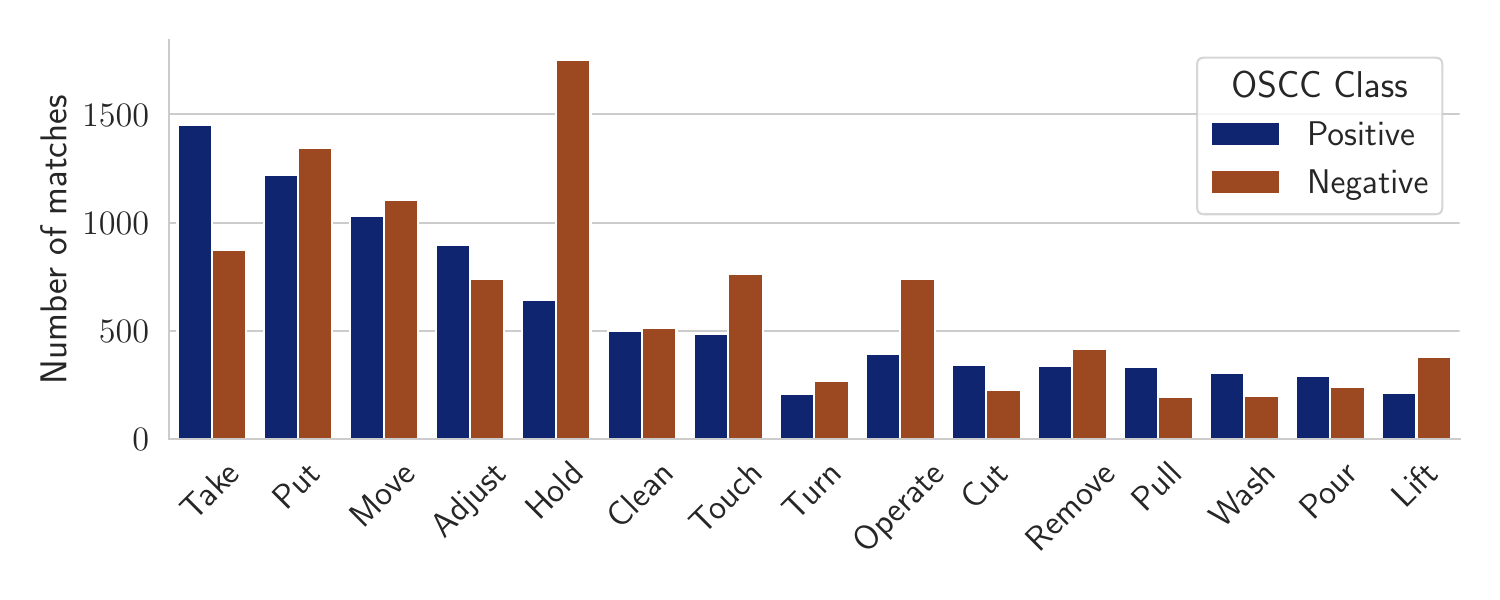}
        \caption{{\footnotesize AR Task Prototypes}}
    \end{subfigure}
    \begin{subfigure}{\columnwidth}
        \includegraphics[width=0.95\columnwidth, trim=0.25cm 0.8cm 0.25cm 0.25cm]{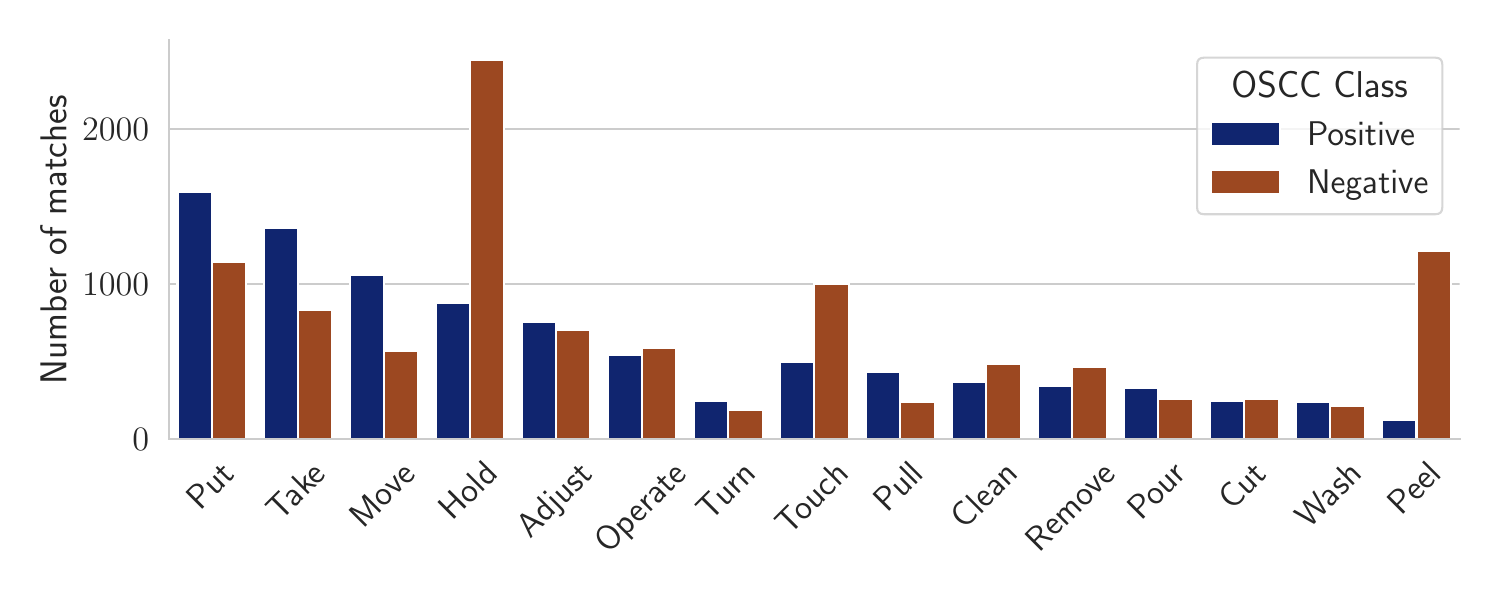}
        \caption{{\footnotesize PNR Task Prototypes}}
    \end{subfigure}
    \caption{Closest nodes to the OSCC samples among AR and PNR task prototypes. Some nodes appear to be more discriminative of the presence or absence of an object state change.}
    \vspace{-0.5cm}
    \label{fig:ossc_matches}
\end{figure}
We evaluate which are the closest task-specific prototypes in Fig.~\ref{fig:ossc_matches}.
In this example, OSCC is the novel task and the model has access to the prototypes of the learnt tasks, namely LTA, AR and PNR.
We focus on the prototypes from the last two tasks and group together nodes that share the same verb label to make the picture more readable.
Looking at the number of occurrences of the prototypes, we observe that some nodes are more discriminative to detect a state change, \eg \textit{peel} and \textit{hold} actions are typically associated (\textit{peel}) or not (\textit{hold}) with state changes, and therefore show more evident peaks for positive and negative classes, indicating the network is using these clues to solve the task.

\paragraph{Confusion matrices}
\begin{figure}
    \centering
    \footnotesize
    \vspace{-.1cm}
    \begin{subfigure}[b]{0.49\columnwidth}
        \centering
        \includegraphics[width=1\columnwidth,trim=0.4cm 0.25cm 0.4cm 0.5cm]{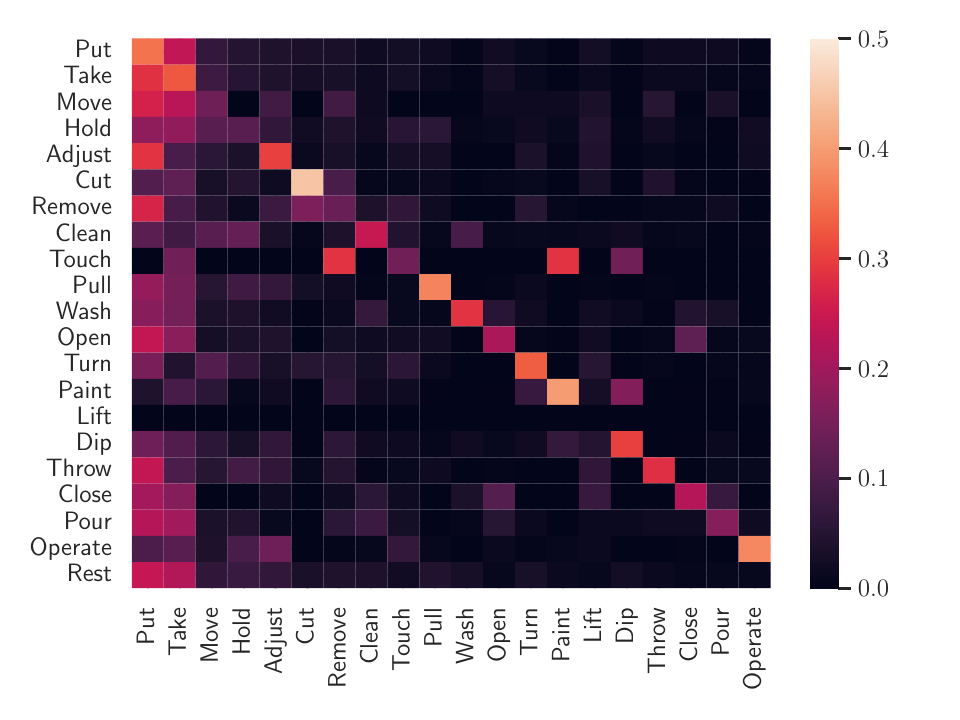}
        \caption{{\footnotesize Verb (\emph{MLP})}}
        \label{fig:verb_baseline}
    \end{subfigure}
    \hfill
    \begin{subfigure}[b]{0.49\columnwidth}  
        \centering 
        \includegraphics[width=1\columnwidth,trim=0.4cm 0.25cm 0.4cm 0.5cm]{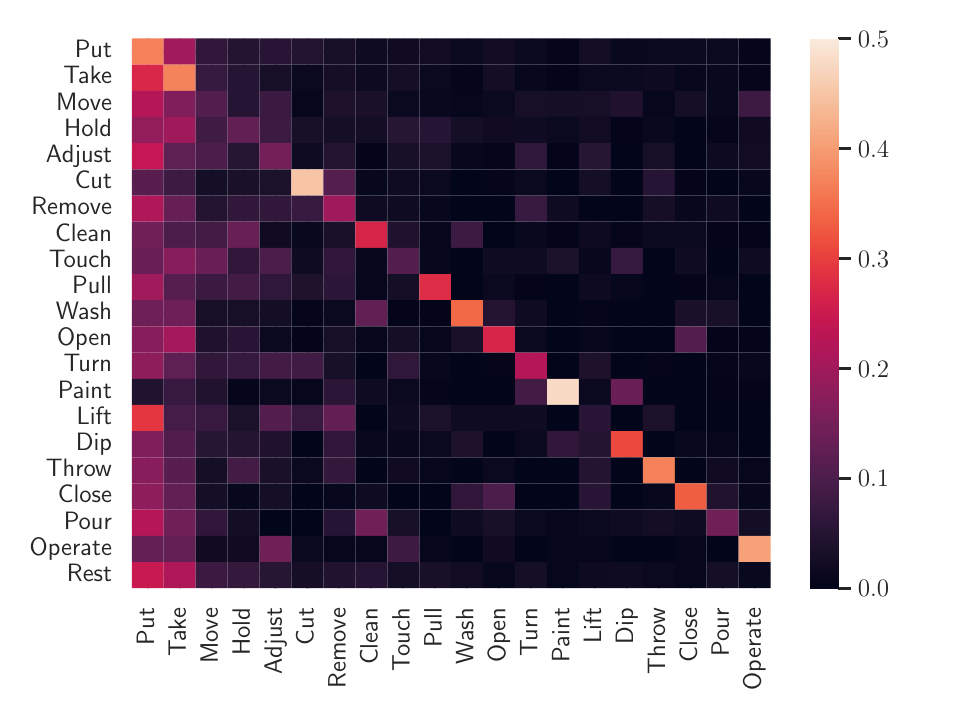}
        \caption{{\footnotesize Verb (\ours)}}
        \label{fig:verb_egopack}
    \end{subfigure}
    \vskip\baselineskip
    \begin{subfigure}[b]{0.49\columnwidth}   
        \centering 
        \includegraphics[width=1\columnwidth,trim=0.75cm 0.25cm 0.75cm 0.5cm]{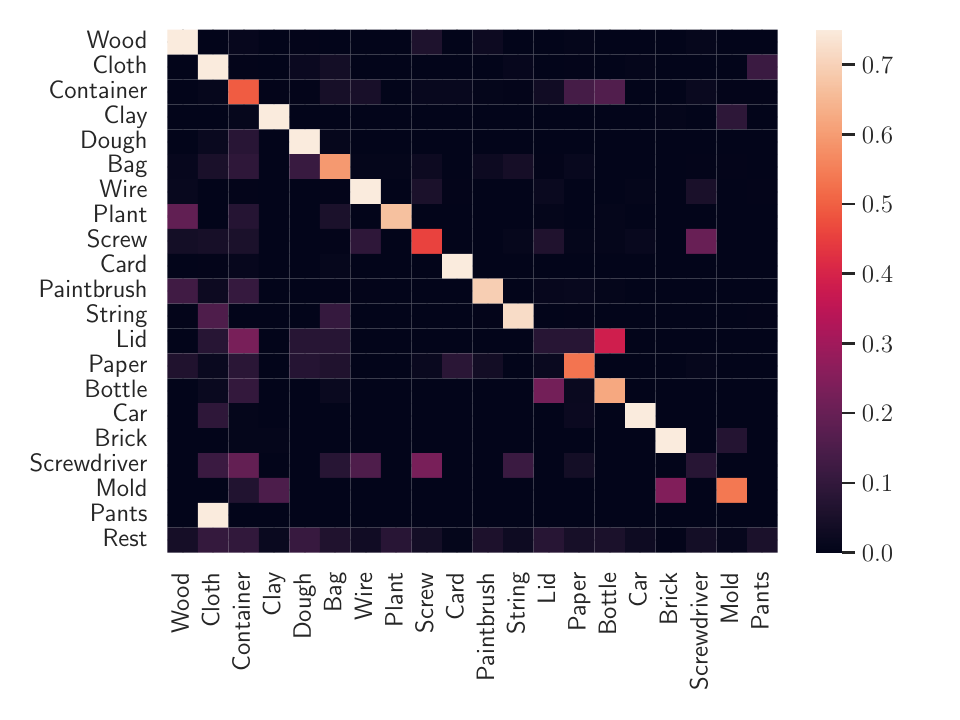}
        \caption{{\footnotesize Noun (\emph{MLP})}}    
        \label{fig:noun_baseline}
    \end{subfigure}
    \hfill
    \begin{subfigure}[b]{0.49\columnwidth}   
        \centering 
        \includegraphics[width=1\columnwidth,trim=0.75cm 0.25cm 0.75cm 0.5cm]{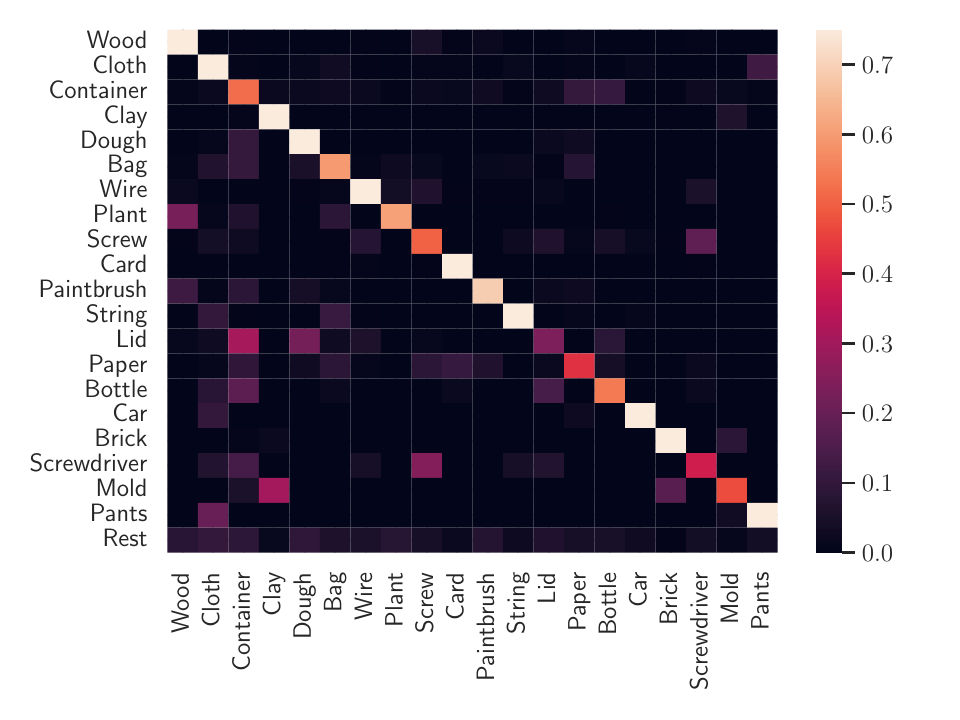}
        \caption{{\footnotesize Noun (\ours)}} 
        \label{fig:noun_egopack}
    \end{subfigure}
    \caption{Action Recognition confusion matrix of \ours compared to the MLP \textit{baseline} for the top-20 verb and noun classes.} 
    \vspace{-0.5cm}
    \label{fig:cm}
\end{figure}
In Fig.~\ref{fig:cm}, we compare the confusion matrix of \ours with the MLP model for the top-20 largest verb and noun classes in the AR task, grouping the remaining classes in a \quotes{\textit{rest}} pseudo-class.
Overall, we observe an evident improvement on the noun labels, due to the positive effect of cross-tasks interaction. 
For example, the network appears to better disambiguate between objects that may appear at the same time in the scene, \eg \quotes{\textit{pants}} and \quotes{\textit{cloth}} or \quotes{\textit{bottle}} and \quotes{\textit{lid}}, which we speculate to be the result of a better ability of other tasks, namely OSCC, to identify active objects.
Regarding the verbs, we also observe notable improvements, in addition to better recognition of verbs that are the temporal inverse of each other, \eg \quotes{\textit{put}} and \quotes{\textit{take}} or \quotes{\textit{open}} and \quotes{\textit{close}}, thanks to the improved temporal reasoning of our unified model.

\section{Conclusions and future work}
\label{sec:conclusions}
We presented \ours, a framework that allows knowledge sharing between different egocentric vision tasks, enabling an efficient use of the perspectives that each task can provide.
We built \ours on top of a unified temporal architecture that can model distinct tasks with a shared backbone and minimal task-specific overhead.
\ours overcomes the main limitation posed by traditional multi-task learning approaches, namely the unrealistic expectation that supervision is available for all tasks at training time. 
Indeed, the prototypes mechanism behind \ours allows to create a summary of what the model has learnt so far as it abstracts the task-specific knowledge that could be used in novel tasks.
The model can then be updated to the any new task, while also peeking at the perspective of the previous tasks.
Results on \egofourd validate our approach, showing competitive performance with other methods, even without end to end training.

\section*{Acknowledgements}
This study was carried out within the FAIR - Future Artificial Intelligence Research and received funding from the European Union Next-GenerationEU (PIANO NAZIONALE DI RIPRESA E RESILIENZA (PNRR) – MISSIONE 4 COMPONENTE 2, INVESTIMENTO 1.3 – D.D. 1555 11/10/2022, PE00000013). This manuscript reflects only the authors’ views and opinions, neither the European Union nor the European Commission can be considered responsible for them. We acknowledge the CINECA award under the ISCRA initiative, for the availability of high performance computing resources and support.

{
    \small
    \bibliographystyle{ieeenat_fullname}
    \bibliography{main}
}

\section*{Appendix}
\setcounter{section}{0}
\renewcommand{\thesection}{\Alph{section}}


In Appendix~\ref{supp:clarification_temporal} we provide additional details on how the different tasks of \egofourd~\cite{ego4d} are modelled as a temporal graph, expanding Sec.~\ref{sec:experiments_impl_details} of the main paper. 
Appendix~\ref{supp:task_translation} provides additional implementation details of the \emph{Task Translation} model and how the EgoT2 architecture was adapted to our scenario.
Appendix~\ref{supp:mtl} presents additional experiments to evaluate the role of negative transfer in MTL.
A more in-depth comparison of the methods on the test-set of the \egofourd challenges is shown in Appendix~\ref{supp:test_set}.
Finally, we show more qualitative results in Appendix~\ref{supp:qualitatives}.

\begin{table*}[htbp]
\centering
\footnotesize
\begin{tabular}{lcccccc}
\toprule

& \multicolumn{2}{c}{\textbf{AR}} & \multicolumn{1}{c}{\textbf{OSCC}} & \multicolumn{2}{c}{\textbf{LTA}} & \multicolumn{1}{c}{\textbf{PNR}} \\
\cmidrule(lr){2-7}

& \textbf{Verbs Top-1 (\%)} & \textbf{Nouns Top-1 (\%)} & \textbf{Acc. (\%)} & \textbf{Verbs ED ($\downarrow$)} & \textbf{Nouns ED ($\downarrow$)} & \textbf{Loc. Err. (s) ($\downarrow$)}\\

\midrule

Temporal Graph & \textbf{24.25} & \textbf{30.43} & \textbf{71.26} & 0.754 & 0.752 & \textbf{0.61} \\

\midrule

Multi-Task Learning & 22.16 & 29.34 & 70.93 & 0.740 & \textbf{0.746} & 0.62 \\

Multi-Task Learning (+ PCGrad~\cite{gradient_surgery}) & 22.01 & 29.46 & 70.86 & \textbf{0.737} & \textbf{0.746} & 0.63 \\

\bottomrule

\end{tabular}
\caption{Results of PCGrad~\cite{gradient_surgery} compared to vanilla Multi-Task Learning.}
\label{tab:supp_mtl}
\end{table*}
\begin{table*}[htbp]
\centering
\footnotesize
\begin{tabular}{lccccc}

\toprule

\multirow{2}{*}{\textbf{\shortstack{\\PNR}}} & \multirow{2}{*}{\textbf{\shortstack{Pre-trained\\ on \egofourd~\cite{ego4d}}}} & \multirow{2}{*}{\textbf{\shortstack{Trained on\\pre-extracted features}}} & \multicolumn{3}{c}{\multirow{2}{*}{\textbf{Loc. Error (s) ($\downarrow$)}}}\\
& & & & & \\

\midrule

CNN LSTM~\cite{ego4d} & \xmark & \xmark & \multicolumn{3}{c}{0.76} \\

EgoVLP~\cite{lin2022egocentric} & \cmark & \xmark & \multicolumn{3}{c}{0.67} \\

EgoT2~\cite{egot2} & \xmark & \xmark & \multicolumn{3}{c}{\textbf{0.66}} \\

\ours & \xmark & \cmark & \multicolumn{3}{c}{\textbf{0.66}} \\

\midrule
\midrule

\multirow{2}{*}{\textbf{\shortstack{\\OSCC}}} & \multirow{2}{*}{\textbf{\shortstack{Pre-trained\\ on \egofourd~\cite{ego4d}}}} & \multirow{2}{*}{\textbf{\shortstack{Trained on\\pre-extracted features}}} & \multicolumn{3}{c}{\multirow{2}{*}{\textbf{Accuracy (\%)}}}\\
& & & & & \\

\midrule

I3D RN-50~\cite{ego4d} & \xmark & \xmark & \multicolumn{3}{c}{67.6} \\

EgoVLP~\cite{lin2022egocentric} & \cmark & \xmark & \multicolumn{3}{c}{74.0} \\

EgoT2 (EgoVLP)~\cite{egot2} & \cmark & \xmark & \multicolumn{3}{c}{\textbf{75.0}} \\
EgoT2 (I3D)~\cite{egot2} & \xmark & \xmark & \multicolumn{3}{c}{71.0} \\

\ours (SlowFast) & \xmark & \cmark & \multicolumn{3}{c}{72.1} \\

\midrule
\midrule

\multirow{2}{*}{\textbf{\shortstack{\\LTA}}} & \multirow{2}{*}{\textbf{\shortstack{Pre-trained\\ on \egofourd~\cite{ego4d}}}} & \multirow{2}{*}{\textbf{\shortstack{Trained on\\pre-extracted features}}} & \multirow{2}{*}{\textbf{Verb ($\downarrow$)}} & \multirow{2}{*}{\textbf{Noun ($\downarrow$)}} & \multirow{2}{*}{\textbf{Action ($\downarrow$)}} \\
& & & & &\\

\midrule

SlowFast~\cite{ego4d} & \xmark & \xmark & 0.739 & 0.780 & 0.943 \\

EgoT2~\cite{egot2} & \xmark & \xmark & 0.722 & 0.764 & 0.935 \\

HierVL~\cite{hiervl} & \cmark & \xmark & 0.724 & \textbf{0.735} & 0.928 \\

I-CVAE~\cite{mascaro2023intention} & \xmark & \cmark & 0.741 & 0.740 & 0.930 \\

\ours & \xmark & \cmark & \textbf{0.721} & \textbf{0.735} & \textbf{0.925} \\

\bottomrule

\end{tabular}
\caption{
Comparison of \ours on the test set of the Ego4D benchmarks, highlighting differences in terms of additional \egofourd pretraining and use of pre-extracted features.
}
\label{tab:supp_challenge_details}
\end{table*}
\begin{figure*}[htbp]
    \centering
    \footnotesize
    \vspace{.2cm}
    \begin{subfigure}[b]{0.32\textwidth}
        \centering
        \includegraphics[width=1\columnwidth,trim=0.4cm 0cm 0.4cm 0.5cm]{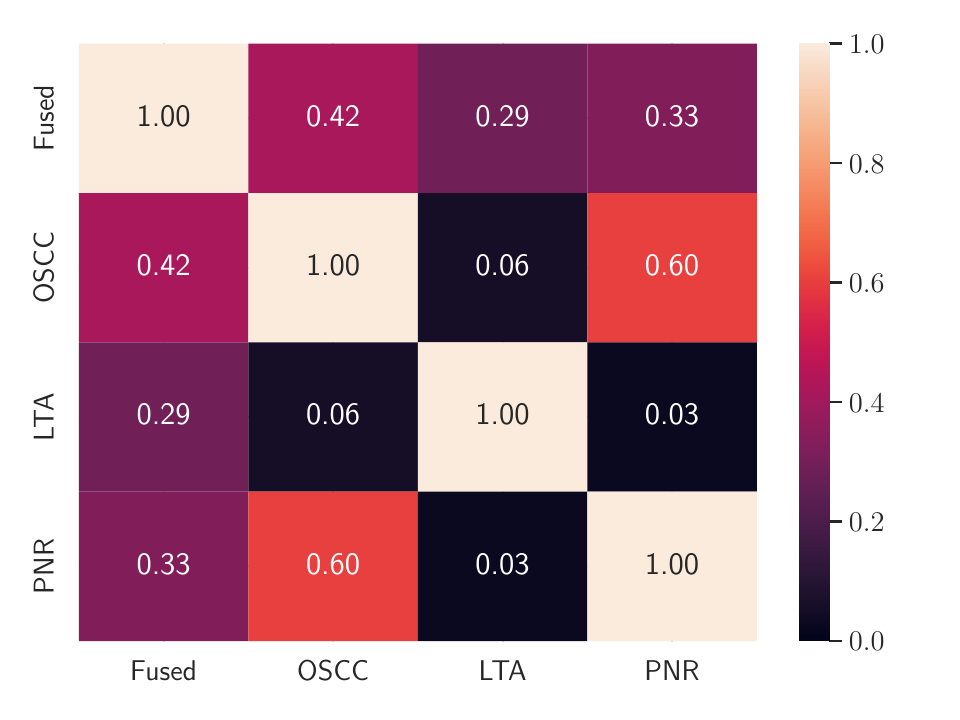}
        \caption{{\footnotesize AR Verb}}
        \label{fig:aggr_ar_verb}
    \end{subfigure}
    \hfill
    \begin{subfigure}[b]{0.32\textwidth}
        \centering 
        \includegraphics[width=1\columnwidth,trim=0.4cm 0cm 0.4cm 0.5cm]{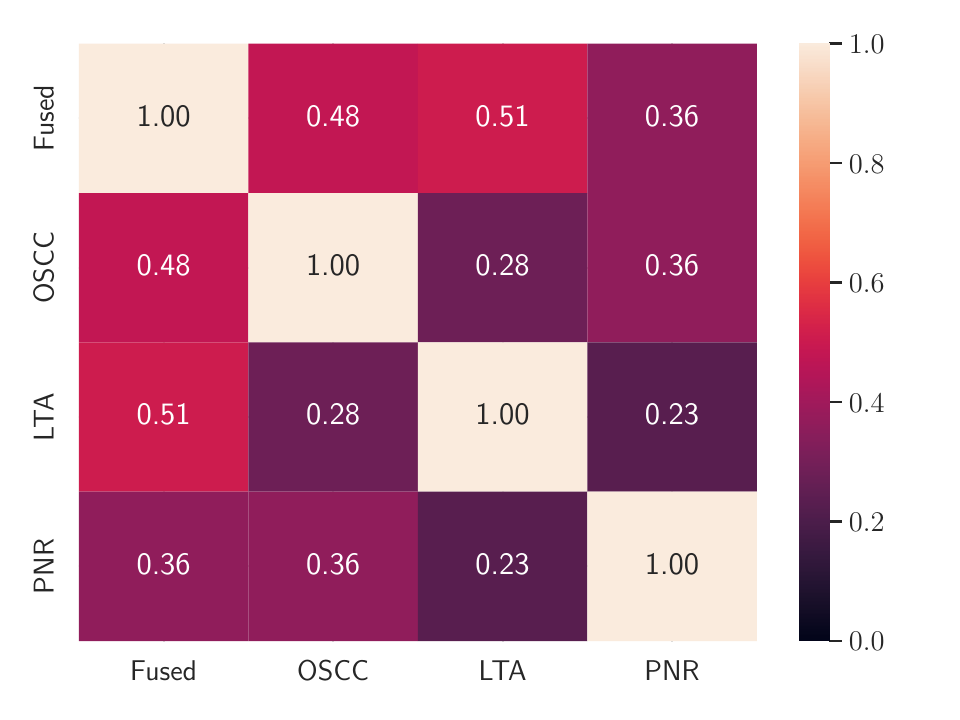}
        \caption{{\footnotesize AR Noun}}
        \label{fig:aggr_ar_noun}
    \end{subfigure}
    \hfill
    \begin{subfigure}[b]{0.32\textwidth}
        \centering 
        \includegraphics[width=1\columnwidth,trim=0.4cm 0cm 0.4cm 0.5cm]{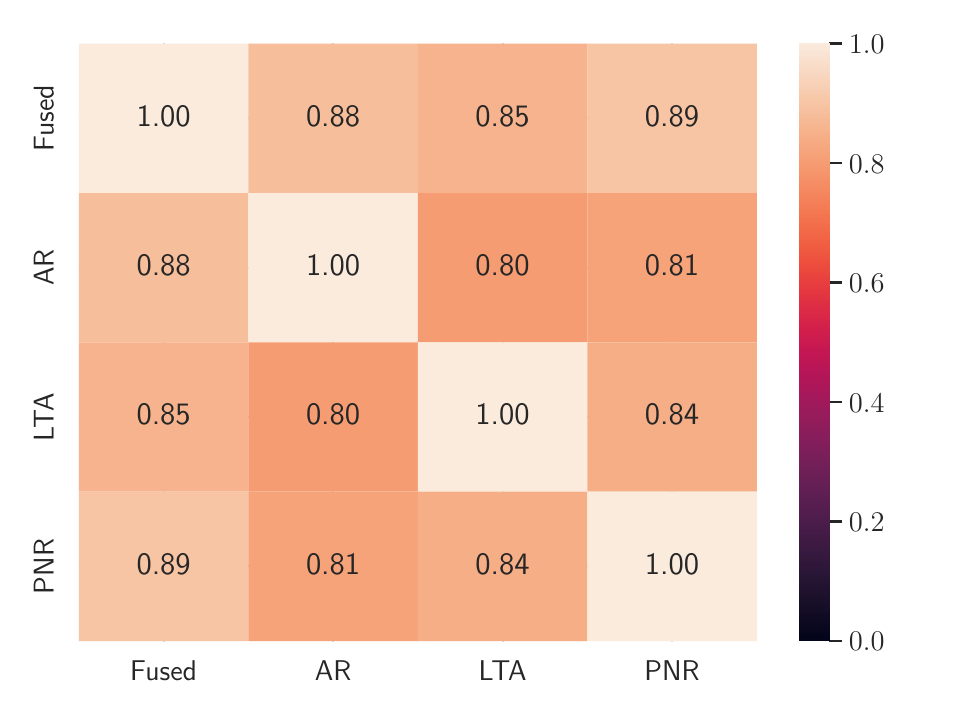}
        \caption{{\footnotesize OSCC}}    
        \label{fig:aggr_oscc}
    \end{subfigure}
    \caption{Agreement ratio between predictions from different tasks when the novel task is Action Recognition (Fig.~\ref{fig:aggr_ar_verb} and Fig.~\ref{fig:aggr_ar_noun}) and Object State Change Classification (Fig.~\ref{fig:aggr_oscc}). \emph{Fused} represents the sum of the logits from the auxiliary tasks.} 
    \label{fig:aggreements}
\end{figure*}

\section{Additional implementation details}\label{supp:clarification_temporal}

\subsection{Temporal modelling}
\paragraph{Action Recognition (AR)}
Action annotations are derived from the LTA benchmark.
Each action is an annotated segment lasting approx. $8.0$ seconds, and actions may temporally overlap. To provide each action with additional temporal context from the surrounding, we process actions in fixed length sequences of $w=9$ actions, each mapped to a node in the temporal graph.
The target action is the central node of the sequence and the classification loss is computed only on this node. 
The window size was selected to be larger than the receptive field of the temporal GNN after 3 layers of graph convolution, which we observed to be the optimal depth of the network. Furthermore, using fixed size sequences allows to train the model on videos containing a variable and possibly large number of actions.

\paragraph{Long Term Anticipation (LTA)}
LTA is formulated as an action anticipation task in which the model is shown $N=2$ input clips and has to predict the actions occurring in the following $Z=20$ timestamps. As in AR, clips lasts approx. $8.0$ seconds and may overlap. Therefore, the effective temporal window seen by the model may vary between $8.0$ and $16.0$ seconds, depending on how much the input clips overlap.
Input clips and future timestamps are mapped to nodes in the graph, with the latter initialised with the mean of the features of the input clips. 
Edges connect each node to its subsequent and preceding nodes. 
Additionally, nodes that represent future actions to be predicted are also connected to the input clips. 
This connectivity pattern allows local temporal reasoning, \eg to rearrange the order of actions in the anticipation window, while using the global context provided by the input clips to guide the prediction. Similarly to AR, there is a \textit{one-to-one} correspondence between actions and nodes of the graph.

\paragraph{Object State Change Classification (OSCC) and Point of No Return (PNR)}
Unlike action-based annotations, OSCC and PNR do not necessarily match the boundaries of an action segment of the video. Each segment lasts approx. $8.0$ seconds and is uniformly divided in $4$ (OSCC) or $16$ (PNR) smaller sub-segments that are mapped to the nodes of the graph.

\subsubsection{Temporal model sharing across tasks}
A key premise of \ours is that different tasks are modelled using the same shared temporal backbone architecture, even though the temporal granularity of the different tasks may vary. 
To achieve this, we do not constraint nodes to represent the same fixed size temporal window across all tasks. 
Through the utilisation of a multi-task learning process, we force the network to jointly learn tasks with different temporal resolutions, enabling reasoning at different temporal scales. 
This formulation is particularly effective to prepare the model to new tasks, as the model has already learnt to combine tasks with potentially different temporal resolutions during the MTL training. 
As an example, consider the case in which the MTL model is trained on AR, LTA, OSCC and PNR. 
In this case, the nodes of the temporal graph represent actions when the task is AR or LTA, or shorter temporal sub-segments for OSCC and PNR. 
To train \ours, we update the weights of $\mathcal{M}_t$ for all novel tasks, except LTA for which we observe better performance by not updating the temporal model. 

\section{Task Translation Implementation Details}\label{supp:task_translation}

The objective of the \emph{Task Translation} experiments is to compare the task translation mechanism proposed by Ego-T2~\cite{egot2}, which learns a mapping between features extracted from different task-specific models, to \ours which leverages past gained knowledge under the form of task-specific prototypes. 
For fair comparison, we re-implement this mechanism and evaluate it on top of the same temporal backbone and the same pre-extracted features of \ours. 
We start from the EgoT2-g model and employ the same architecture for the \emph{Task Translation}, which consists of a 1-layer encoder-decoder stack, each with 8 heads, dropout 0.1 and features size 1024. 
The input of the \emph{Task Translation} is provided by an ensemble of \emph{Temporal Graph} models, one for each task.
The whole architecture is trained for one task at the time, as \ours, and only the encoder-decoder architecture is updated, while the temporal models that compose the ensemble are kept frozen.
We train \emph{Task Translation} for 30 epochs, using the Adam optimiser with learning rate $1\times 10^{-4}$ (with the exception of OSCC which uses learning rate $1\times 10^{-3}$), batch size $16$, linear warmup for the first 5 epochs and weight decay $1\times 10^{-5}$.

\section{Additional Multi-Task Experiments}\label{supp:mtl}
\begin{table*}[ht]
\centering
\footnotesize
\begin{tabular}{lcccccc}
\toprule
& \multicolumn{2}{c}{\textbf{AR}} & \multicolumn{1}{c}{\textbf{OSCC}} & \multicolumn{2}{c}{\textbf{LTA}} & \multicolumn{1}{c}{\textbf{PNR}} \\
\cmidrule(lr){2-7}
& \textbf{Verbs Top-1 (\%)} & \textbf{Nouns Top-1 (\%)} & \textbf{Acc. (\%)} & \textbf{Verbs ED ($\downarrow$)} & \textbf{Nouns ED ($\downarrow$)} & \textbf{Loc. Err. (s) ($\downarrow$)}\\
\midrule
MTL (All tasks) & 22.05 & 29.44 & 71.10 & 0.740 & \textbf{0.746} & 0.62 \\
MTL (3 tasks) + FT & 24.36 & \textbf{31.31} & 71.60 & 0.744 & 0.754 & 0.62 \\
MTL (3 tasks) + TT & 22.30 & 29.50 & 70.96 & 0.738 & 0.757 & 0.62 \\
\midrule
\ours & \textbf{25.10} & 31.10 & \textbf{71.83} & \textbf{0.728} & 0.752 & \textbf{0.61} \\
\bottomrule
\end{tabular}
\caption{Comparison of vanilla MTL and two finetuning strategies to extend MTL models to novel tasks.}
\label{tab:supp_mtl_ft}
\end{table*}

MTL suffers from negative transfers between different tasks, and fine-tuning an MTL on a new task may not be the most effective approach to retain knowledge learned in the MTL training process.
We observe evidence of this phenomenon in Table~\ref{tab:supp_mtl_ft}, where we compare the MTL on all tasks with two finetuning approaches to extend a model trained on three tasks to a fourth novel task. 
MTL+FT finetunes the model for the novel task, as already shown in Table~\ref{tab:ablation_tasknomy}, while MTL+TT replaces the EgoPack’s second stage with a decoder analogous to TT, which learns the new task as a \quotes{recombination} of the previous tasks.

\paragraph{Brute Force Multi-Task Learning}
\begin{table*}[htbp]
\centering
\footnotesize
\begin{tabular}{cccccccccc}
\toprule
& & & & \multicolumn{2}{c}{\textbf{AR}} & \multicolumn{1}{c}{\textbf{OSCC}} & \multicolumn{2}{c}{\textbf{LTA}} & \multicolumn{1}{c}{\textbf{PNR}} \\
\cmidrule(lr){5-10}

\textbf{AR} & \textbf{OSCC} & \textbf{LTA} & \textbf{PNR} & \textbf{Verbs Top-1 (\%)} & \textbf{Nouns Top-1 (\%)} & \textbf{Acc. (\%)} & \textbf{Verbs ED ($\downarrow$)} & \textbf{Nouns ED ($\downarrow$)} & \textbf{Loc. Err. (s) ($\downarrow$)}\\
\midrule
\multicolumn{4}{c}{Single tasks} & 24.25 & 30.43 & 71.26 & 0.754 & 0.752 & \textbf{0.61} \\
\midrule
\cmark & \cmark & - & - & 23.98 & 30.60 & 70.81 & - & - & - \\ 
\cmark & - & \cmark & - & 22.23 & 29.48 & - & 0.744 & 0.744 & - \\ 
\cmark & - & - & \cmark & 24.05 & 30.72 & - & - & - & 0.63 \\ 
- & \cmark & \cmark & - & - & - & 70.71 & 0.745 & 0.751 & - \\ 
- & \cmark & - & \cmark & - & - & 71.01 & - & - & 0.66 \\ 
- & - & \cmark & \cmark & - & - & - & 0.751 & 0.752 & 0.62\\ 
\midrule
\cmark & \cmark & \cmark & - & 22.05 & 29.44 & 71.10 & 0.739 & 0.745 & - \\ 
\cmark & \cmark & - & \cmark & 23.82 & 30.83 & 71.03 & - & - & 0.63 \\ 
\cmark & - & \cmark & \cmark & 22.24 & 29.83 & - & 0.745 & \textbf{0.743} & 0.62 \\ 
- & \cmark & \cmark & \cmark & - & - & 71.06 & 0.746 & 0.751 & 0.63 \\ 
\midrule
\multicolumn{4}{c}{MTL (All tasks)} & 22.05 & 29.44 & 71.10 & 0.740 & 0.746 & 0.62 \\ 
\midrule
\multicolumn{4}{c}{\textbf{\ours}} & \textbf{25.10} & \textbf{31.10} & \textbf{71.83} & \textbf{0.728} & 0.752 & \textbf{0.61}\\ 
\bottomrule
\end{tabular}
\caption{Brute force experiments in multi-task learning with all combinations of tasks.}
\label{tab:supp_brute_force}
\end{table*}
Table~\ref{tab:supp_brute_force} presents a comprehensive analysis of MTL on all task combinations, to assess the effect of negative transfer when a smaller subset of tasks is used. 
Even with fewer tasks, MTL still suffers from negative transfer across tasks and does not represent an upper bound for \ours, which is showing a clear advantage.

\paragraph{Minimising negative transfer}
Various approaches have been proposed to address the issue of negative transfer in multi-task learning~\cite{kendall2018multi,chen2018gradnorm,sinha2018gradient,guo2018dynamic,vandenhende2020mti,gradient_surgery}.
Although the multi-task setting significantly differs from the settings proposed for \ours, we provide a comparison with one of these methods, PCGrad~\cite{gradient_surgery}, which projects tasks' gradients on the normal plane of all the other gradients to remove interference among tasks. 
Apart from minimal fluctuations, PCGrad does not appear to significantly improve over MTL, showing that these methods may still be insufficient to effectively reduce the negative transfer, as shown in Table~\ref{tab:supp_mtl}.

\section{Comparison of methods on the test-set}\label{supp:test_set}
We summarise the main differences between \ours and the other methods on the test-set in Table~\ref{tab:supp_challenge_details}, extending Table~\ref{tab:challenge} of the main paper and highlighting differences in terms of additional \egofourd pretraining and use of pre-extracted features.
\ours relies on pre-extracted features from Omnivore~\cite{omnivore}, which was trained on Kinetics-400~\cite{quo_vadis} for action recognition. 
As a result, these features are highly semantic and may struggle to encode finer temporal details required by certain tasks, \eg to detect changes in the objects being manipulated in OSCC or PNR.
Most other methods, with the exception of I-CVAE~\cite{mascaro2023intention}, train also their features extraction backbones on \egofourd benchmarks' data, which allows to learn task-specific models more suited for the task at hand.
On the contrary, we do not update the features extraction backbone when training \ours.

When evaluating \ours on the test-set, we also observe a significant performance gap compared to other methods that rely on some amount of additional data from \egofourd, while the benchmarks data are more limited in size. HierVL~\cite{hiervl} is pretrained on the full Ego4D using a contrastive video-language objective with short-term and long-term textual narrations. EgoVLP~\cite{lin2022egocentric} is pretrained on a large subset of Ego4D using a video-language contrastive objective with action-aware positive samples and scene-aware negative samples. 
The only method directly comparable to \ours in terms of pre-training data and parameters updated is I-CVAE~\cite{mascaro2023intention}, which uses the SlowFast~\cite{slowfast} features released by~\cite{ego4d} for the LTA benchmark.
The extension of \ours to additional backbones, possibly with end-to-end finetuning, is outside of the scope of this paper and is left as a future work.
For OSCC, we report the results of \ours using SlowFast features instead of Omnivore as they showed better performances compared to the latter.

\section{Additional qualitative results}\label{supp:qualitatives}
\ours fuses the predictions coming from different task perspectives by summing the task-specific logits.
We show in Fig.~\ref{fig:aggreements} the agreement ratio between the predictions produced by the different tasks $\mathbf{y}_{i}^{k}$ and the final output computed as the sum of the individual contributions $\mathbf{y}_{i} = \sum_k \mathbf{y}_{i}^{k} $.
In Action Recognition, we observe low agreement both between task pairs and with respect to the fused predictions, suggesting that they contribute complementary information to the novel task. 
On the other hand, in OSCC, tasks predictions tend to be more consistent across tasks.

\end{document}